\documentclass{article}
\usepackage{iclr2027_conference,times}

\usepackage{amsmath,amsfonts,bm}

\def\eqref#1{equation~\ref{#1}}

\def\1{\bm{1}}

\DeclareMathAlphabet{\mathsfit}{\encodingdefault}{\sfdefault}{m}{sl}
\SetMathAlphabet{\mathsfit}{bold}{\encodingdefault}{\sfdefault}{bx}{n}

\usepackage{amsmath,amssymb}
\usepackage{booktabs}
\usepackage{flafter}
\usepackage{float}
\usepackage{graphicx}
\usepackage{hyperref}
\usepackage{microtype}
\usepackage{url}
\usepackage{xspace}
\usepackage{makecell}
\usepackage{multirow}
\usepackage[table]{xcolor}
\hypersetup{hidelinks}
\usepackage{enumitem}
\usepackage{placeins}
\usepackage{booktabs}
\usepackage{diagbox}
\usepackage{caption}

\title{TempoKV: Predictive KV Cache Quantization for Autoregressive Video Generation}

\author{%
Jiahui Hu\textsuperscript{1,2}\thanks{Equal contribution.}\quad
Weile Mo\textsuperscript{1}\footnotemark[1]\quad
Peiliang Cai\textsuperscript{1}\quad
Jiacheng Liu\textsuperscript{1}\\
\bfseries Jinkui Ren\textsuperscript{3}\quad
Xiantao Zhang\textsuperscript{3}\quad
Xiaobing Tu\textsuperscript{3}\quad
Linfeng Zhang\textsuperscript{1}\thanks{Corresponding author.}\\[3pt]
\normalfont\small
\textsuperscript{1}\,Shanghai Jiao Tong University\quad
\textsuperscript{2}\,South China University of Technology\quad
\textsuperscript{3}\,Alibaba Group
}
\hypersetup{%
  pdftitle={TempoKV: Predictive KV Cache Quantization for Autoregressive Video Generation},
  pdfauthor={Jiahui Hu, Weile Mo, Peiliang Cai, Jiacheng Liu, Jinkui Ren, Xiantao Zhang, Xiaobing Tu, Linfeng Zhang}
}

\newcommand{\method}{TempoKV\xspace}

\iclrfinalcopy

\begin{document}
\maketitle
\lhead{Preprint.}

\vspace{-1em}
\begin{abstract}
Autoregressive (AR) video diffusion models enable streaming generation of long videos 
but suffer from significant memory overhead due to the rapidly growing KV cache. 
Existing KV cache quantization methods are primarily designed for LLMs and transfer poorly to video models, where activations exhibit different statistics and strong spatiotemporal redundancy.
We observe that the video KV cache is largely predictable over time, with a clear asymmetry. Keys are highly predictable from their temporal history, while values depend primarily on the current key and need a short history of preceding values. 
Based on these observations, we propose \textbf{\method}, a training-free predictive KV cache quantization framework for AR video generation. \method predicts keys and values according to their distinct dependencies and stores only the residuals at low precision. Both predictors are lightweight, calibrated once offline by linear regression on a small set of BF16 samples, and remain fixed across prompts and generation runs. Experiments on Self-Forcing, Causal Forcing, and LongCat-Video-13B show that \method consistently achieves higher generation quality than existing KV cache quantization methods at comparable or higher compression ratios, reaching up to $6.71\times$ KV cache compression with the best overall quality--efficiency tradeoff. 
Our code will be released on GitHub.

\end{abstract}

\section{Introduction}
\label{sec:introduction}


Recent video diffusion models~\citep{wan2025wanopenadvancedlargescale, wu2025hunyuanvideo15technicalreport} have achieved high generation quality, but extending generation to long-duration videos still requires significant computation and memory.
Autoregressive approaches address this limitation by generating videos progressively while reusing previously generated context
\citep{huang2025selfforcing,zhu2026causalforcing,liu2025rollingforcing}. 
However, the resulting \emph{KV cache} becomes another bottleneck, as its size grows linearly with generation duration and quickly dominates GPU memory.
For example, in Causal Forcing~\citep{zhu2026causalforcing}, generating a 45-second 480p video requires 51.76\,GB solely for the KV cache when stored in BF16, exceeding the memory of most consumer GPUs and limiting further extension of the generation horizon.

Existing KV cache quantization methods~\citep{liu2024kivi,ashkboos2024quarot, kang2024gear, hooper2024kvquant} are developed primarily for large language models (LLMs), focusing on the numerical characteristics of language KV caches. Autoregressive video diffusion models, however, exhibit different activation distributions and strong spatiotemporal redundancy across consecutive latent frames, making the direct transfer of these methods to video models suboptimal.

\begin{figure}[H]
    \centering
    \includegraphics[width=\linewidth]{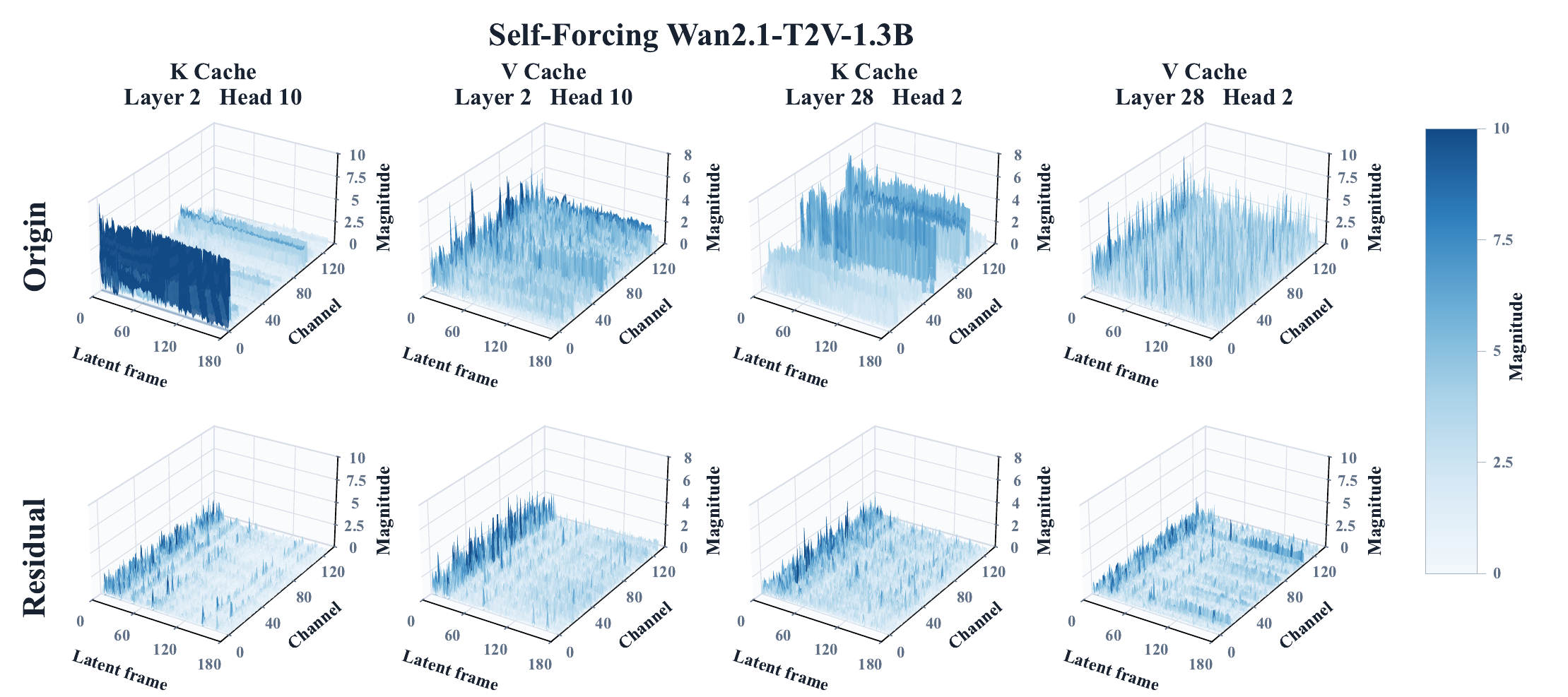}
    \caption{\textbf{ Magnitudes of KV caches and their prediction residuals on Self-Forcing.} Top: the original KV cache magnitudes at representative layers and heads. Bottom: the corresponding residual magnitudes after applying \method. Prediction reduces the dynamic range of both keys and values, suppressing outliers across latent frames and channels. }
    \label{fig:kv_magnitude}
\end{figure}

Recent work, QVG~\citep{xi2026quantvideogen}, takes an important step toward video-specific quantization by grouping similar video tokens and subtracting group centroid. Notably, QVG observes different compression behavior between keys (K) and values (V): the same smoothing operation reduces the quantization error of keys much more than that of values. Despite this discrepancy, it applies the same similarity-based formula to both. This difference suggests that video KV redundancy is not fully captured by token similarity alone. 
To further investigate, we analyze how the KV cache changes over latent frames and whether this evolution is predictable. As shown in the top row of Fig.~\ref{fig:kv_magnitude}, keys evolve smoothly and predictably along the latent frames, while values are markedly more irregular. 

\vspace{-2em}
We find a clear asymmetry between keys and values. The current key can be accurately predicted from preceding keys at the same spatial position: a linear probe on the key history reaches a cosine similarity of 0.877. Values show a different pattern. Previous values alone are far less predictive of the current value, reaching a cosine similarity of only 0.637, which rises to 0.881 once the aligned current key is included, i.e., the key at the same latent frame and spatial position.

Motivated by this observation, we introduce \textbf{\method}, a predictive KV cache quantization framework for autoregressive video generation. \method first predicts the key $K_t$ from reconstructed historical keys, then quantizes only its residual. After reconstructing the current key, \method uses the reconstructed key $\widehat{K}_t$ and reconstructed historical values to predict the current value $V_t$, and again quantizes the remaining residual.
As shown in Fig.~\ref{fig:kv_magnitude}, prediction suppresses prominent outliers and  narrows the dynamic range, producing low-magnitude residuals that are easier to quantize.

This asymmetry has two further implications for quantization. First, the aligned current K provides the primary predictive signal for V, while additional V history offers only limited benefit. In contrast, K relies mainly on its own temporal history and benefits from a longer context. This suggests using a shorter temporal context for V than for K. Second, the asymmetry also affects precision allocation. Quantization error in K not only degrades its own reconstruction, but is also carried into the subsequent V prediction through the cross predictor. This makes K more sensitive to error. Accordingly, we assign K a higher precision than V except at the lowest-precision INT2 setting.

Experiments on Self-Forcing~\citep{huang2025selfforcing}, Causal Forcing~\citep{zhu2026causalforcing}, and LongCat-Video-13B~\citep{longcat2025video} validate the asymmetric predictive structure and demonstrate its effectiveness for low-bit KV compression. \method improves generation quality over existing low-bit KV cache quantization methods at comparable or higher compression ratios, reaching up to $\mathbf{6.71\times}$. These gains remain robust across diverse autoregressive generation frameworks and resolutions, indicating that the KV asymmetry is not specific to a single backbone.

In summary, our main contributions are:
\begin{itemize}[leftmargin=*, labelsep=0.5em, itemsep=5pt, parsep=0pt, topsep=0pt]
    \item \textbf{Asymmetric Predictive Dependencies.}
    We identify that K and V exhibit asymmetric predictive dependencies across latent frames. Keys are highly predictable from their history, whereas values benefit from the aligned current key in addition to previous values. Through layer-wise and temporal analysis, we show that this asymmetry is consistent across layers and throughout generation.

    \item \textbf{\method Framework.}
    Inspired by this asymmetry, we propose \method, a training-free predictive quantization framework for autoregressive video diffusion models. \method predicts keys from their temporal history and values through a key-to-value cross predictor, assigns higher precision to keys, and quantizes only the remaining residuals. The predictors are fitted offline by linear regression on short BF16 traces with minimal overhead.
    
    \item \textbf{Strong Quality--Efficiency Tradeoff.}
    We evaluate \method across diverse architectures and configurations. \method outperforms prior low-bit KV cache quantization methods on all fidelity metrics, achieving competitive or higher compression with low runtime overhead.
\end{itemize}

\section{Related Work}
\label{sec:related_work}

\subsection{Autoregressive Video Generation}

Recent autoregressive video models generate extended sequences causally while reusing previously generated context.
To bridge bidirectional diffusion and causal generation, CausVid~\citep{yin2024causvid} adapts pretrained bidirectional diffusion models into autoregressive generators through distribution matching distillation, while Self-Forcing~\citep{huang2025selfforcing} reduces exposure bias by training on autoregressive rollouts with self-generated outputs.
Self-Forcing++~\citep{cui2025selfforcingpp} further mitigates long-horizon error accumulation using teacher guidance on self-generated videos.
Other works extend long-horizon generation through joint denoising and attention
sinks~\citep{liu2025rollingforcing}, frame-level autoregression with short-window
attention~\citep{yang2025longlive}, chunk-wise prediction~\citep{sandai2025magi1},
frame-context packing~\citep{zhang2025framepack}, or explicit memory
mechanisms~\citep{xiao2025worldmem,henschel2024streamingt2v}.
LongCat-Video-13B ~\citep{longcat2025video} adopts video continuation pretraining with coarse-to-fine generation.
However, these AR approaches rely on
historical context. Consequently, the linearly growing KV cache rapidly dominates GPU memory, becoming a severe bottleneck for high-resolution and long-video generation.

\subsection{KV Cache Eviction and Compaction}

To mitigate the growing KV-cache memory bottleneck, early efforts in LLMs selectively retain important context, such as attention sinks together with a recent window~\citep{xiao2023streamingllm}, recent and heavy-hitter tokens~\citep{zhang2023h2o}, or clustered important KV positions identified from an observation window~\citep{li2024snapkv}. Recent video-specific methods adapt cache management to autoregressive video. 
PaFu-KV~\citep{chen2026pafukv} models time-heterogeneous token importance with a Salience Estimation Head distilled from a bidirectional teacher, 
while FutureForcing~\citep{luo2026futureforcing} uses future query proxies to guide token eviction and merging. Others use heterogeneous memory allocation (DummyForcing~\citep{guo2026dummyforcing}) or hybrid static and dynamic pruning (Forcing-KV~\citep{ji2026forcingkv}), while PackCache~\citep{li2026packcache} dynamically compacts the cache through condition anchoring and cross-frame decay modeling.

\subsection{KV Cache Quantization}
LLM KV-cache quantization spans several directions. KIVI~\citep{liu2024kivi} and KVQuant~\citep{hooper2024kvquant} adapt quantization to the distinct distributions of KV through asymmetric quantization granularity and non-uniform quantization. QuaRot~\citep{ashkboos2024quarot} and RotateKV~\citep{su2025rotatekv} use rotations to mitigate the effect of outliers, while SKVQ~\citep{duanmu2024skvq} and ZipCache~\citep{he2024zipcache} retain recent or salient tokens at higher precision. PQCache~\citep{zhang2024pqcache} and CommVQ~\citep{li2025commvq} further explore product and vector quantization. Despite this diversity, these methods target language-model KV statistics. QVG~\citep{xi2026quantvideogen} extends low-bit KV quantization to autoregressive video generation by leveraging spatiotemporal redundancy with Semantic-Aware Smoothing, followed by coarse-to-fine Progressive Residual Quantization. \method takes a complementary predictive view: rather than grouping similar tokens, it models the temporal dependencies of K and V and uses their asymmetry to guide the quantization design.

\section{Method}
\label{sec:method}
\begin{figure*}[t]
    \centering
    \includegraphics[width=\textwidth]{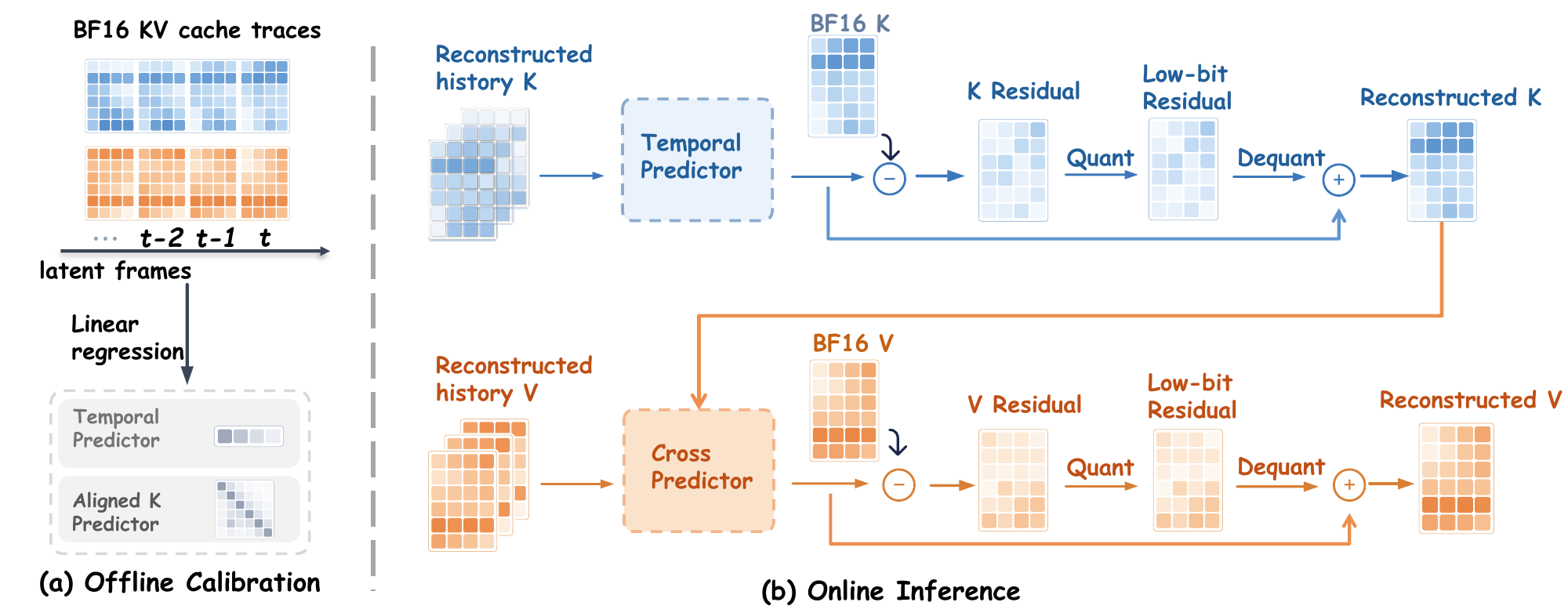}

    \caption{\textbf{\method Framework.} (a) Offline Calibration: a clean BF16 KV trace is collected from 8 calibration prompts, and predictors are fitted on it by linear regression.  (b) Online Inference: $\widetilde{K}_{t,s}$ is
    predicted from the reconstructed K history, and $\widetilde{V}_{t,s}$ from the
    aligned $\widehat{K}_{t,s}$ together with the reconstructed V history. Only the
    low-bit residual $R_{t,s}$ is stored.}
    \label{fig:tempokv_overview}
    \vspace{-1em}
\end{figure*}

\subsection{Preliminaries}
\label{sec:preliminaries}

An autoregressive video model generates a sequence of $T$ latent frames, each containing $S$ spatial tokens with consistent spatial indexing across time. For a given attention layer and head, let $K_{t,s},\,V_{t,s}\in\mathbb{R}^{D}$ denote the key and value at spatial position $s$ in latent frame $t$, where $D$ is the head dimension.
We use standard round-to-nearest quantization.
For a tensor $X$, its $b$-bit affine quantization and reconstruction are defined as
\begin{equation}
    Q_b(X)
    =
    \left\lfloor
        \frac{X}{s_X}
    \right\rceil + z_X,
    \qquad
    \hat{X}
    =
    \bigl(Q_b(X)-z_X\bigr)s_X,
    \label{eq:quantizer}
\end{equation}
where $s_X=(\max(X)-\min(X))/(2^b-1)$ and
$z_X=\lfloor-\min(X)/s_X\rceil$.

\subsection{Asymmetric Predictability of Video KV Caches}
\label{sec:asymmetry}

KV caches are  highly correlated across the latent frames. 
To further examine this dependency and characterize how it differs between K and V, we fit four linear probes and measure the cosine similarity between the predicted and original caches. The first predicts $K_{t,s}$ from the key history at the same position, and the second predicts $V_{t,s}$ from the value history. The other two add one more input, the other cache of the current frame: $V_{t,s}$ for K prediction and $K_{t,s}$ for V prediction.

\begin{figure}[t]
\centering
    \includegraphics[width=\linewidth]{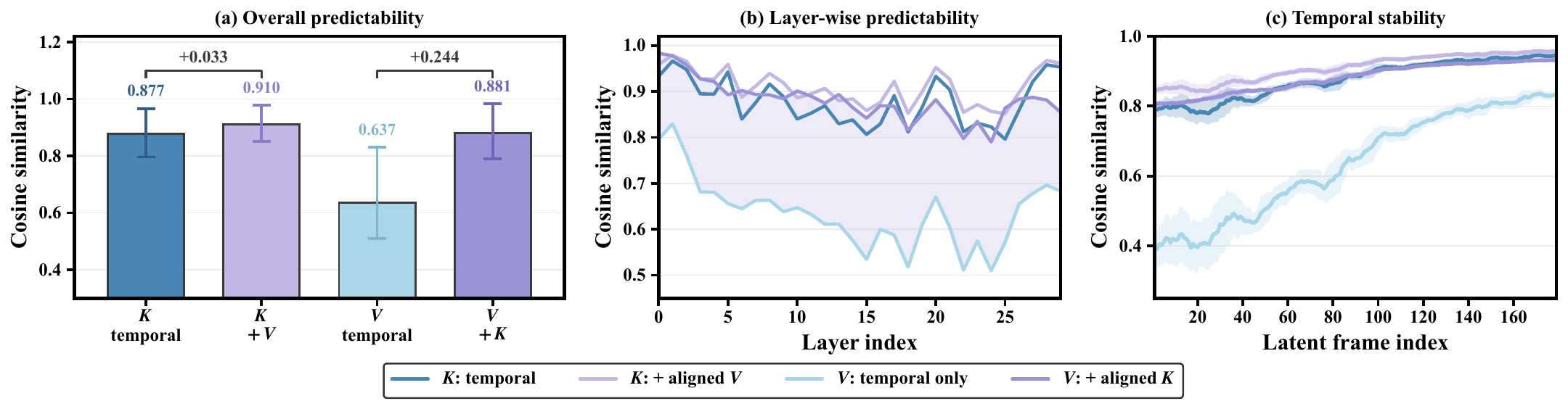}
    \caption{\textbf{Asymmetric predictability of video KV cache.}
    (a) Temporal history predicts keys well, while values benefit substantially from the aligned key.
    (b) This advantage holds across all layers.
    (c) The gain remains stable over long generation.}
    \vspace{-1em}
\label{fig:asymmetry}
\end{figure}

Fig.~\ref{fig:asymmetry} reveals a clear asymmetry between K and V. 
The key history predicts $K_{t,s}$ with a cosine similarity of \textbf{0.877}. The value history predicts $V_{t,s}$ with only $0.637$. Adding $V_{t,s}$ moves the K probe from $0.877$ to $0.910$, a gain of $0.033$. Adding $K_{t,s}$ moves the V probe from $0.637$ to \textbf{0.881}, a gain of \textbf{0.244}. The key history alone already provides an accurate estimate of the current key, while the value history is
considerably less predictive of the current value.
The same trend is observed in each layer and at every frame. In Fig.~\ref{fig:asymmetry}(b), $K_{t,s}$ improves V prediction in all 30 layers, by $0.148$ to $0.307$. The reverse gain, from $V_{t,s}$ to K, ranges from $0.009$ to $0.056$. Fig.~\ref{fig:asymmetry}(c) follows both probes along the latent frames. Both grow more accurate as generation proceeds, and the change is larger for V. The gain from $K_{t,s}$ to V remains higher than the reverse gain throughout the trajectory.

This asymmetry determines the direction of our cross predictor. 
We first predict $K_{t,s}$ from the key history,
then predict $V_{t,s}$ from the reconstructed key
$\hat{K}_{t,s}$ and the value history. The two cross directions cannot be used simultaneously at decoding, since each one needs the other cache to be available first. We keep the direction $K\!\rightarrow\!V$ that provides much larger gain.

Keys and values also require different precision. Key quantization perturbs the attention logits before softmax and can alter the resulting attention weights. Value quantization, in contrast, affects the attention output only after the weights are formed. This asymmetry is a property of attention. The cross predictor introduces a second and independent source of asymmetry.
The quantization error in K not only affects the reconstructed key itself but also propagates to the value prediction via the cross predictor.
Therefore, we assign higher precision to keys than values.

The observation further motivates different temporal history lengths. K is predicted from the key history alone and keeps improving as that history grows. V already receives a strong input from the current key, and extra value history adds little. A longer history also increases predictor parameters and keeps more data in the cache during decoding. Thus, we use $h_K=2$ and $h_V=1$ by default.

\subsection{Predictive Residual Quantization}

\label{sec:predictive_quantization}
\paragraph{\method framework.}

\method predicts the KV cache from its own dependencies and quantizes only the residual (Fig.~\ref{fig:tempokv_overview}). Let $\widetilde{X}_{t,s}$ and $\widehat{X}_{t,s}$ denote the prediction and reconstruction, respectively.
\begin{equation}
    R^X_{t,s}=X_{t,s}-\widetilde{X}_{t,s},
    \qquad
    \widehat{X}_{t,s}
    =
    \widetilde{X}_{t,s}
    +
    D_b\!\left(Q_b(R^X_{t,s})\right).
    \label{eq:residual_quant}
\end{equation}

Following Sec.~\ref{sec:asymmetry}, \method uses asymmetric predictive dependencies for K and V. Let $o_t^K\leq h_K$ and $o_t^V\leq h_V$ denote the available key and value history lengths, respectively. We use $\beta^K$ and $\beta^V$ to denote the bias terms of the K and V predictors. The key is predicted from its reconstructed history:
\begin{equation}
    \widetilde{K}_{t,s}
    =
    \beta^K
    +
    \sum_{i=1}^{o_t^K}
    A^K_{i}
    \odot
    \widehat{K}_{t-i,s},
    \label{eq:k_predictor}
\end{equation}
where $\odot$ denotes element-wise multiplication. For the first frames after the anchor, fewer than $h_K$, we use a separate set of coefficients. After reconstructing the current key from its quantized prediction residual, the value is predicted from both $\widehat{K}_{t,s}$ and its reconstructed history. A dense projection $W_{K\rightarrow V}\in\mathbb{R}^{D\times D}$ maps the aligned key to the value:
\begin{equation}
    \widetilde{V}_{t,s}
    =
    \beta^V
    +
    W_{K\rightarrow V}\widehat{K}_{t,s}
    +
    \sum_{j=1}^{o_t^V}
    A^V_j
    \odot
    \widehat{V}_{t-j,s}.
    \label{eq:v_predictor}
\end{equation}

\paragraph{Anchor frames.}
\method uses only the first latent frame as an anchor throughout the entire generation process.
Unlike residual frames, the anchor is quantized without prediction. Its KV activations retain a larger dynamic range than the prediction residuals, making aggressive low-bit quantization more prone to large errors. Since anchor initializes subsequent reconstruction, preserving its accuracy is particularly important. We quantize anchor at a higher precision, $b_a=\max(4,b_K,b_V)$, where $b_K$ and $b_V$ denote the bit-widths of the K and V residuals. 

\paragraph{Offline Calibration.}Both predictors are fitted offline on BF16 keys and values recorded from a small set of calibration prompts. The current full-precision key or value serves as the regression target, while the predictor inputs are the corresponding caches
after quantization and reconstruction:

\begin{equation}
    \theta_X^{\star}
    =
    \arg\min_{\theta_X}
    \sum_{t,s}
    \bigl\|
        X_{t,s}-f_X\bigl(\mathcal{C}^{X}_{t,s};\theta_X\bigr)
    \bigr\|_2^2
    +
    \lambda_X\|\theta_X\|_2^2,
    \qquad
    X\in\{K,V\},
    \label{eq:calibration}
\end{equation}

where $f_X$ is the predictor of Eq.~\ref{eq:k_predictor}
or~\ref{eq:v_predictor}, $\theta_K=\{A^K, \beta^K\}$ and
$\theta_V=\{A^V, \beta^V, W_{K\rightarrow V}\}$ collect its coefficients, $\mathcal{C}^{K}_{t,s}$ is the reconstructed key history, and $\mathcal{C}^{V}_{t,s}$ the reconstructed aligned key together with the reconstructed value history.
We first fit the key predictor and use the resulting reconstructed keys to fit the value predictor, following the asymmetric dependency used during inference. The recorded BF16 traces are shared across compression
configurations. Different bit-width settings are obtained by applying the
corresponding quantization and reconstruction to the same traces, without
rerunning the model. Predictor fitting is performed entirely on
these recorded traces, requiring neither backpropagation through the model nor retraining. Further details are provided in Appendix~\ref{app:calibration}.

\section{Experiments}
\label{sec:experiments}

\subsection{Settings}
\label{sec:setup}

\paragraph{Models and baselines.}
We evaluate \method on three autoregressive video generation models: Self-Forcing-Wan-1.3B ~\citep{huang2025selfforcing}, Causal Forcing-Wan-1.3B ~\citep{zhu2026causalforcing}, and LongCat-Video-13B~\citep{longcat2025video}. We compare against three groups of baselines: (1)video-specific KV cache quantization, represented by QVG~\citep{xi2026quantvideogen} and QVG-Pro, (2)general-purpose KV cache quantization, including RTN, KIVI~\citep{liu2024kivi}, and QuaRot~\citep{ashkboos2024quarot}. We evaluate all video generation on the first 128 prompts from MovieGen ~\citep{polyak2024moviegen}, using VBench-Long ~\citep{huang2023vbench,huang2025vbench++}
metrics to assess perceptual video quality.
Visual Quality is the average of the six reported VBench dimensions.


\vspace{0.9em}
\begin{figure}[H]
    \centering
    \vspace{-1em}
    \includegraphics[
        width=\textwidth
    ]{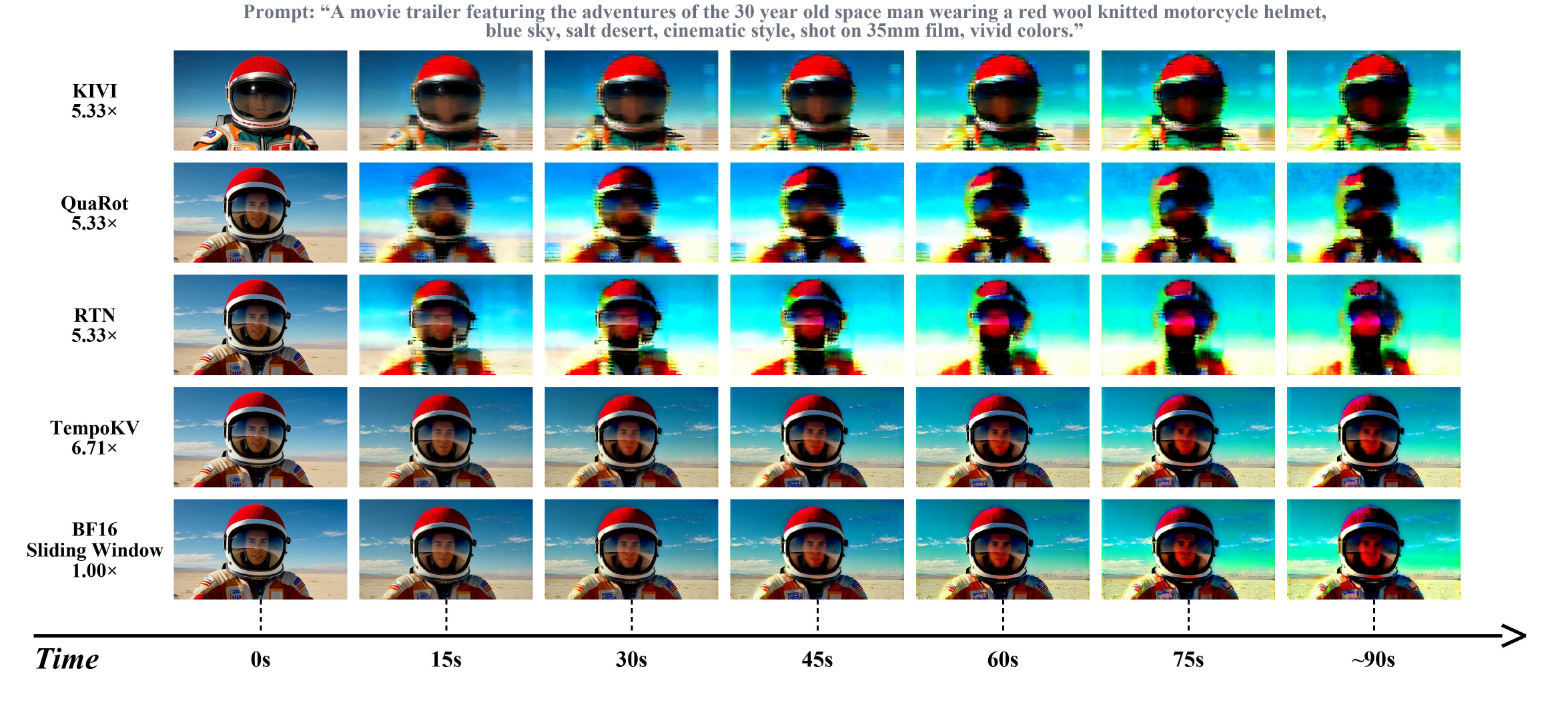}
    \caption{Qualitative comparison of long-horizon generation with Self-Forcing. While KIVI, QuaRot,
    and RTN exhibit progressively stronger visual degradation over time, \method maintains greater temporal consistency, with results remaining close to the BF16 reference.}
    \label{fig:long_horizon_qualitative}
\end{figure}
\vspace{-1em}

\begin{table}[H]
\caption{Causal Forcing results on MovieGen at $832\times480$ and 717 frames.
Best and second-best results within each precision setting are shown in \textbf{bold} and \underline{underline}, respectively.}
\vspace{-1em}
\label{tab:main_results_cf}

\centering
\scriptsize
\setlength{\tabcolsep}{2.2pt}

\resizebox{\textwidth}{!}{%
\begin{tabular}{l|c|cccc|cccccc|c}

\toprule
\makecell[c]{\textbf{Method}}
&
\makecell[c]{\textbf{Compression}\\\textbf{Ratio}}
&
\makecell[c]{\textbf{PSNR} $\uparrow$}
&
\makecell[c]{\textbf{SSIM} $\uparrow$}
&
\makecell[c]{\textbf{LPIPS} $\downarrow$}
&
\makecell[c]{\textbf{FID} $\downarrow$}
&
\makecell[l]{\textbf{Background}\\\textbf{Consistency} $\uparrow$}
&
\makecell[l]{\textbf{Subject}\\\textbf{Consistency} $\uparrow$}
&
\makecell[l]{\textbf{Dynamic}\\\textbf{Degree} $\uparrow$}
&
\makecell[l]{\textbf{Imaging}\\\textbf{Quality} $\uparrow$}
&
\makecell[l]{\textbf{Aesthetic}\\\textbf{Quality} $\uparrow$}
&
\makecell[l]{\textbf{Motion}\\\textbf{Smoothness} $\uparrow$}
&
\makecell[l]{\textbf{Visual}\\\textbf{Quality} $\uparrow$}
\\
\midrule

BF16
& 1.00$\times$
& $\infty$
& 1.00
& 0.00
& 0.00
& 95.91
& 95.81
& 58.25
& 61.18
& 53.46
& 97.08
& 76.95 \\

\midrule
\multicolumn{13}{c}{\textit{INT2 KV Cache, 480p}} \\
\midrule

RTN
& 5.33$\times$
& 10.32
& 0.42
& 0.58
& 143.05
& 95.76
& 95.05
& 15.05
& 42.05
& 41.22
& 96.73
& 64.31 \\

KIVI
& 5.33$\times$
& 10.57
& 0.33
& 0.66
& 168.80
& 95.29
& 95.09
& 12.43
& 42.75
& 43.03
& \underline{97.43}
& 64.34 \\

QuaRot
& 5.33$\times$
& 9.51
& 0.38
& 0.64
& 174.07
& 95.83
& 94.18
& 39.77
& 36.35
& 49.48
& 95.83
& 68.57 \\

QVG
& \underline{6.61$\times$}
& 16.21
& 0.70
& 0.22
& 14.24
& 95.99
& 95.78
& 54.62
& 58.58
& 52.42
& \textbf{97.47}
& 75.81 \\

QVG-Pro
& 4.34$\times$
& 16.51
& 0.71
& 0.20
& 9.67
& 95.98
& \underline{95.94}
& 57.95
& \underline{61.00}
& \textbf{53.34}
& 97.42
& 76.94 \\

\rowcolor{blue!8}
\textbf{\method(\texttt{k2v2})}
& \textbf{6.71$\times$}
& \underline{17.23}
& \underline{0.76}
& \underline{0.17}
& \underline{9.15}
& \underline{96.01}
& 95.80
& \textbf{59.21}
& 60.84
& \underline{53.28}
& 97.11
& \underline{77.04} \\

\rowcolor{blue!15}
\textbf{\method(\texttt{k4v2})}
& 4.73$\times$
& \textbf{22.85}
& \textbf{0.79}
& \textbf{0.13}
& \textbf{6.34}
& \textbf{96.03}
& \textbf{95.96}
& \underline{59.16}
& \textbf{61.12}
& 53.27
& 97.15
& \textbf{77.11}
\\

\midrule
\multicolumn{13}{c}{\textit{INT4 KV Cache, 480p}} \\
\midrule

RTN
& 3.20$\times$
& 14.48
& 0.66
& 0.27
& 15.86
& 95.94
& 95.72
& 54.45
& 59.69
& 52.66
& 97.15
& 75.94 \\

KIVI
& 3.20$\times$
& 14.50
& 0.63
& 0.29
& 46.50
& 95.24
& 95.00
& 30.50
& 52.66
& 50.52
& 97.66
& 70.26 \\

QuaRot
& 3.20$\times$
& 14.61
& 0.67
& 0.26
& 14.36
& 95.90
& 95.85
& 56.52
& 60.08
& 52.55
& 97.29
& 76.37 \\

QVG
& \underline{3.08$\times$}
& 16.44
& 0.71
& 0.20
& 9.39
& 95.97
& 95.91
& 58.80
& 61.03
& \underline{53.34}
& 97.38
& 77.07 \\

QVG-Pro
& 2.51$\times$
& 16.38
& 0.71
& 0.20
& 9.20
& 95.98
& 95.96
& 58.19
& \underline{61.16}
& 53.30
& \underline{97.47}
& 77.01 \\

\rowcolor{blue!8}
\textbf{\method(\texttt{k5v3})}
& \textbf{3.67$\times$}
& \underline{23.21}
& \underline{0.80}
& \underline{0.11}
& \underline{4.03}
& \underline{96.04}
& \underline{95.99}
& \underline{59.00}
& 61.09
& 53.31
& 97.30
& \underline{77.12} \\

\rowcolor{blue!15}
\textbf{\method(\texttt{k6v4})}
& 2.97$\times$
& \textbf{23.85}
& \textbf{0.83}
& \textbf{0.10}
& \textbf{3.75}
& \textbf{96.23}
& \textbf{96.16}
& \textbf{59.06}
& \textbf{61.26}
& \textbf{53.42}
& \textbf{97.77}
& \textbf{77.31}
\\

\bottomrule
\end{tabular}%
}
\end{table}

\begin{table*}[t]
\caption{Self-Forcing results on MovieGen at $832\times480$ and 717 frames.
Best and second-best results within each precision setting are shown in
\textbf{bold} and \underline{underline}, respectively.}
\label{tab:main_results_sf}
\vspace{-0.8em}
\centering
\scriptsize
\setlength{\tabcolsep}{3.0pt}

\resizebox{\textwidth}{!}{%
\begin{tabular}{l|c|cccc|cccccc|c}

\toprule
\makecell[c]{\textbf{Method}}
&
\makecell[c]{\textbf{Compression}\\\textbf{Ratio}}
&
\makecell[c]{\textbf{PSNR} $\uparrow$}
&
\makecell[c]{\textbf{SSIM} $\uparrow$}
&
\makecell[c]{\textbf{LPIPS} $\downarrow$}
&
\makecell[c]{\textbf{FID} $\downarrow$}
&
\makecell[l]{\textbf{Background}\\\textbf{Consistency} $\uparrow$}
&
\makecell[l]{\textbf{Subject}\\\textbf{Consistency} $\uparrow$}
&
\makecell[l]{\textbf{Dynamic}\\\textbf{Degree} $\uparrow$}
&
\makecell[l]{\textbf{Imaging}\\\textbf{Quality} $\uparrow$}
&
\makecell[l]{\textbf{Aesthetic}\\\textbf{Quality} $\uparrow$}
&
\makecell[l]{\textbf{Motion}\\\textbf{Smoothness} $\uparrow$}
&
\makecell[l]{\textbf{Visual}\\\textbf{Quality} $\uparrow$}
\\
\midrule

BF16
& 1.00$\times$
& $\infty$
& 1.00
& 0.00
& 0.00
& 96.36
& 97.88
& 42.39
& 70.39
& 39.77
& 98.61
& 74.23 \\

\midrule
\multicolumn{13}{c}{\textit{INT2 KV Cache, 480p}} \\
\midrule

RTN
& 5.33$\times$
& 12.03
& 0.48
& 0.57
& 227.40
& 95.83 & 96.71 & 12.32 & 43.36 & 38.15 & 98.04 & 64.06 \\

KIVI
& 5.31$\times$
& 14.11
& 0.53
& 0.48
& 150.31
& 95.82 & 97.50 & 12.79 & 51.63 & 38.64 & 98.32 & 65.78 \\

QuaRot
& 5.33$\times$
& 11.68
& 0.34
& 0.60
& 139.98
& 89.07 & 78.92 & 36.34 & 62.76 & 38.57 & 96.12 & 66.96\\

QVG
& \underline{6.61$\times$}
& 18.34
& 0.70
& 0.18
& 15.17
& 95.25
& 97.05
& 39.40
& 66.48
& 39.69
& 98.49
& 72.73 \\

QVG-Pro
& 4.65$\times$
& 18.92
& 0.73
& 0.15
& \underline{9.17}
& 95.34
& \underline{97.15}
& 41.10
& 67.17
& \underline{39.79}
& \underline{98.50}
& 73.18 \\

\rowcolor{blue!8}
\textbf{\method(\texttt{k2v2})}
& \textbf{6.71$\times$}
& \underline{19.85}
& \underline{0.78}
& \underline{0.13}
& 10.04
& \underline{95.35}
& 97.04
& \textbf{41.46}
& \underline{67.81}
& \underline{39.79}
& 98.48
& \underline{73.32} \\

\rowcolor{blue!15}
\textbf{\method(\texttt{k4v2})}
& 4.73$\times$
& \textbf{23.17}
& \textbf{0.85}
& \textbf{0.07}
& \textbf{5.23}
& \textbf{95.83}
& \textbf{97.89}
& \underline{41.37}
& \textbf{70.50}
& \textbf{39.90}
& \textbf{98.73}
& \textbf{74.04} \\

\midrule
\multicolumn{13}{c}{\textit{INT4 KV Cache, 480p}} \\
\midrule

RTN
& 3.20$\times$
& 16.80
& 0.66
& 0.24
& 83.32
& 94.96
& 95.83
& 33.12
& 65.37
& 39.75
& 97.83
& 71.14 \\

KIVI
& 3.19$\times$
& 16.67
& 0.65
& 0.26
& 73.40
& 94.82
& 95.87
& 27.76
& 64.54
& 39.73
& 97.72
& 70.07 \\

QuaRot
& 3.21$\times$
& 16.81
& 0.66
& 0.24
& 47.11
& 94.80
& 96.73
& 36.07
& 66.68
& 39.74
& 97.73
& 71.96 \\

QVG
& \underline{3.49$\times$}
& 18.86
& 0.73
& 0.15
& 9.69
& 95.22
& 96.99
& \underline{41.38}
& 67.15
& 39.76
& 98.46
& 73.16 \\

QVG-Pro
& 3.01$\times$
& 19.21
& 0.74
& 0.14
& 9.30
& 95.33
& 97.12
& 41.34
& 66.96
& \underline{39.80}
& \underline{98.50}
& 73.18 \\

\rowcolor{blue!8}
\textbf{\method(\texttt{k5v3})}
& \textbf{3.65$\times$}
& \underline{24.17}
& \underline{0.88}
& \underline{0.06}
& \underline{4.13}
& \underline{95.79}
& \underline{97.72}
& \textbf{41.43}
& \underline{70.90}
& \textbf{39.89}
& \textbf{98.53}
& \underline{74.04} \\

\rowcolor{blue!15}
\textbf{\method(\texttt{k6v4})}
& 2.97$\times$
& \textbf{24.71}
& \textbf{0.89}
& \textbf{0.05}
& \textbf{3.28}
& \textbf{95.91}
& \textbf{97.95}
& 41.36
& \textbf{70.98}
& \textbf{39.89}
& \textbf{98.53}
& \textbf{74.10} \\

\bottomrule
\end{tabular}%
}
\end{table*}

\begin{table*}[!ht]
\caption{LongCat-Video-13B results on MovieGen at $832\times480$ and 717 frames.
Best and second-best results within each precision setting are shown in \textbf{bold} and \underline{underline}, respectively.}
\label{tab:main_results_longcat}
\vspace{-1em}
\centering
\scriptsize
\setlength{\tabcolsep}{3.0pt}

\resizebox{\textwidth}{!}{%
\begin{tabular}{l|c|cccc|cccccc|c}

\toprule
\makecell[c]{\textbf{Method}}
&
\makecell[c]{\textbf{Compression}\\\textbf{Ratio}}
&
\makecell[c]{\textbf{PSNR} $\uparrow$}
&
\makecell[c]{\textbf{SSIM} $\uparrow$}
&
\makecell[c]{\textbf{LPIPS} $\downarrow$}
&
\makecell[c]{\textbf{FID} $\downarrow$}
&
\makecell[l]{\textbf{Background}\\\textbf{Consistency} $\uparrow$}
&
\makecell[l]{\textbf{Subject}\\\textbf{Consistency} $\uparrow$}
&
\makecell[l]{\textbf{Dynamic}\\\textbf{Degree} $\uparrow$}
&
\makecell[l]{\textbf{Imaging}\\\textbf{Quality} $\uparrow$}
&
\makecell[l]{\textbf{Aesthetic}\\\textbf{Quality} $\uparrow$}
&
\makecell[l]{\textbf{Motion}\\\textbf{Smoothness} $\uparrow$}
&
\makecell[l]{\textbf{Visual}\\\textbf{Quality} $\uparrow$}
\\
\midrule

BF16
& 1.00$\times$
& $\infty$
& 1.00
& 0.00
& 0.00
& 95.70
& 97.77
& 49.33
& 72.45
& 40.09
& 98.53
& 75.65 \\

\midrule
\multicolumn{13}{c}{\textit{INT2 KV Cache, 480p}} \\
\midrule

RTN
& 5.33$\times$
& 14.21
& 0.57
& 0.36
& 96.48
& 95.73 & 97.55 & 30.20 & 56.86 & 38.46 & 97.60 
& 69.40 \\

KIVI
& 5.33$\times$
& 15.33
& 0.63
& 0.34
& 62.02
& 94.96
& 94.39
& 30.67
& 64.07
& 38.84
& 96.63
& 69.93 \\

QuaRot
& 5.33$\times$
& 13.35
& 0.54 
& 0.42 
& 36.48
& 94.58 & 95.18 & 41.00 & 66.91 & 39.56 & 97.20 & 72.40
\\


QVG
& \underline{6.47$\times$}
& 15.75
& 0.66
& 0.25
& 24.29
& 95.61
& 97.66
& 47.33
& 69.54
& 39.95
& 98.52
& 74.77 \\

QVG-Pro
& 4.21$\times$
& 15.88
& 0.67
& 0.20
& 15.00
& \underline{95.78}
& \underline{97.81}
& 46.83
& 71.32
& \textbf{40.10}
& \underline{98.57}
& 75.07 \\

\rowcolor{blue!8}
\textbf{\method(\texttt{k2v2})}
& \textbf{6.66$\times$}
& \underline{16.57}
& \underline{0.71}
& \underline{0.15}
& \underline{12.40}
& \textbf{95.81}
& 97.66
& \underline{48.33}
& \underline{71.62}
& \underline{40.09}
& \textbf{98.72}
& \underline{75.37} \\

\rowcolor{blue!15}
\textbf{\method(\texttt{k4v2})}
& 4.72$\times$
& \textbf{17.39}
& \textbf{0.75}
& \textbf{0.12}
& \textbf{7.13}
& 95.76
& \textbf{97.85}
& \textbf{49.50}
& \textbf{72.32}
& \underline{40.09}
& 98.52
& \textbf{75.67} \\

\midrule
\multicolumn{13}{c}{\textit{INT4 KV Cache, 480p}} \\
\midrule

RTN
& 3.20$\times$
& 14.68
& 0.60
& 0.35
& 40.36
& 95.48
& 97.79
& 47.40
& 66.04
& 38.86
& 97.53
& 73.91 \\

KIVI
& 3.20$\times$
& 16.42
& 0.68
& 0.19
& 16.96
& 95.58
& 97.77
& 46.67
& 72.14
& 40.01
& 98.47
& 75.11
\\

QuaRot
& 3.20$\times$
& 16.64
& 0.71
& 0.16
& 15.37
& \textbf{95.73}
& 97.73
& 46.83
& 72.17
& \underline{40.11}
& 98.48
& 75.18 \\

QVG
& \underline{3.59$\times$}
& 16.82
& 0.73
& 0.14
& 14.63
& 95.63
& 97.79
& \underline{48.50}
& 71.86
& 40.07
& 98.49
& 75.39 \\

QVG-Pro
& 2.75$\times$
& 17.34
& 0.74
& 0.12
& 12.51
& 95.60
& 97.78
& 47.56
& 72.35
& \textbf{40.23}
& \textbf{98.63}
& 75.36 \\

\rowcolor{blue!8}
\textbf{\method(\texttt{k5v3})}
& \textbf{3.62$\times$}
& \underline{18.01}
& \underline{0.78}
& \underline{0.10}
& \underline{6.51}
& 95.59
& \underline{97.83}
& 48.32
& 72.31
& 40.09
& 98.51
& \underline{75.44} \\

\rowcolor{blue!15}
\textbf{\method(\texttt{k6v4})}
& 2.97$\times$
& \textbf{18.42}
& \textbf{0.81}
& \textbf{0.09}
& \textbf{6.24}
& \underline{95.67}
& \textbf{97.87}
& \textbf{48.74}
& \underline{72.38}
& \underline{40.11}
& \underline{98.53}
& \textbf{75.54} \\

\bottomrule
\end{tabular}%
}
\end{table*}

\begin{figure*}[t]
    \centering
    \includegraphics[width=\textwidth]{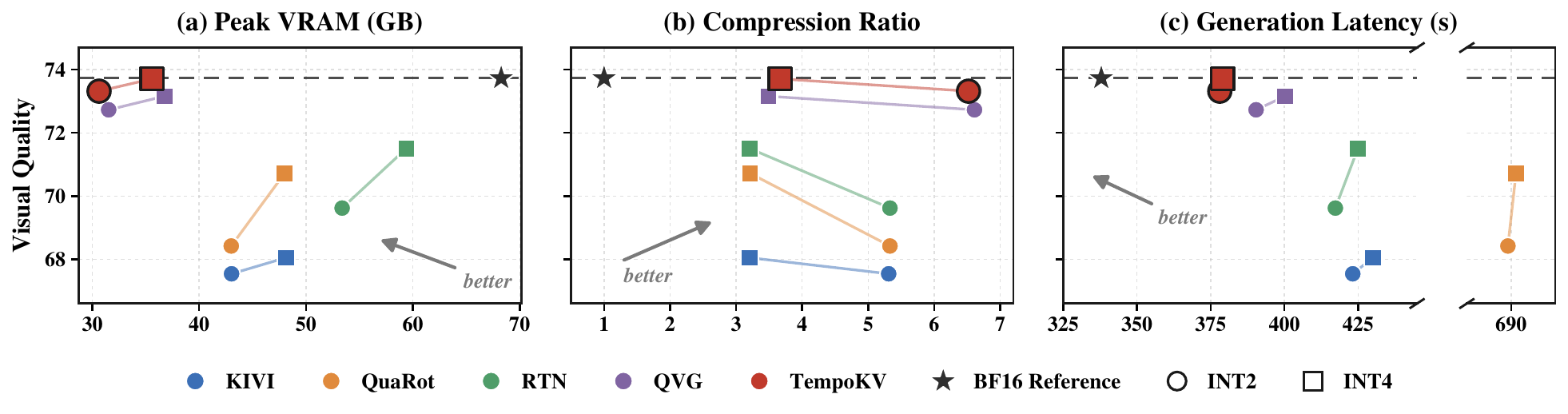}
    \vspace{-1em}
    \caption{
    \textbf{Quality--efficiency tradeoffs.}
    Visual Quality is plotted against \textbf{(a)} peak VRAM, \textbf{(b)} KV-cache compression ratio, and \textbf{(c)} generation latency. \method remains close to BF16 quality while providing the best balance among compression, runtime memory, and generation latency.
    }
    \label{fig:tradeoff}
\end{figure*}

\subsection{Main Results}
\label{sec:main_results}

\paragraph{Quantitative results.}
Tables~\ref{tab:main_results_cf}, \ref{tab:main_results_sf}
and~\ref{tab:main_results_longcat} compare \method with existing methods on three video generation models. Each precision setting includes two \method configurations, where \textbf{\texttt{kNvM}} denotes $N$-bit keys and $M$-bit values.
On Self-Forcing, \method(\textbf{\texttt{k5v3}}) reaches \textbf{$3.65\times$} compression at INT4, exceeding the \textbf{$3.49\times$} of QVG, while improving PSNR from \textbf{18.86} to \textbf{24.17} and Visual Quality from \textbf{73.16} to \textbf{74.04}. 
\method(\textbf{\texttt{k6v4}}) runs at $2.97\times$, close to the $3.01\times$ of QVG-Pro, and improves PSNR from $19.21$ to \textbf{24.71} and FID from $9.30$ to
\textbf{3.28}.
At INT2, \method(\textbf{\texttt{k2v2}}) achieves a higher compression ratio (\textbf{6.71$\times$ vs.\ $6.61\times$}), while improving PSNR from \textbf{18.34} to \textbf{19.85}. 
\method(\textbf{\texttt{k4v2}}) runs at $4.73\times$, close to the $4.65\times$ of QVG-Pro, and
improves PSNR from $18.92$ to \textbf{23.17} and SSIM from 0.73 to \textbf{0.85}.
Consistent gains on Causal Forcing and LongCat-Video-13B further show that these improvements are not specific to a
single backbone or generation setting. 
As shown in Fig.~\ref{fig:long_horizon_qualitative}, \method maintains high visual fidelity and temporal consistency over long-horizon generation, while existing low-bit baselines exhibit severe degradation.

\paragraph{Quality--efficiency tradeoff.}
Fig.~\ref{fig:tradeoff} reports the memory and latency cost of KV cache compression on Self-Forcing. At INT4, \method reduces the KV cache from $48.18$\,GB to \textbf{13.2\,GB} and peak memory usage falls from $68.27$\,GB to \textbf{35.59\,GB}. Under the more aggressive INT2 setting, peak memory usage further decreases to \textbf{30.63\,GB}, while LPIPS remains at \textbf{0.13}. 
Notably, this exceeds the INT4 quality of QVG while using $6.1$\,GB less peak memory. 
Decoding the compressed cache during attention adds latency. On our hardware, \method adds only \textbf{11.9\%} generation overhead to the BF16 latency of $337.9$\,s, against $15.5\%$ for QVG, $23.5\%$ to $25.2\%$ for RTN and KIVI, and $103.9\%$ for
QuaRot. This overhead changes by less than $1$\,s from INT2 to INT4, whereas that of QVG grows by $9.9$\,s.
Together, these results show that \method preserves generation quality while reducing both KV storage and runtime memory, with minimal overhead.

\begin{figure*}[!ht]
    \centering
    \vspace{-1em}
    \includegraphics[width=\textwidth]{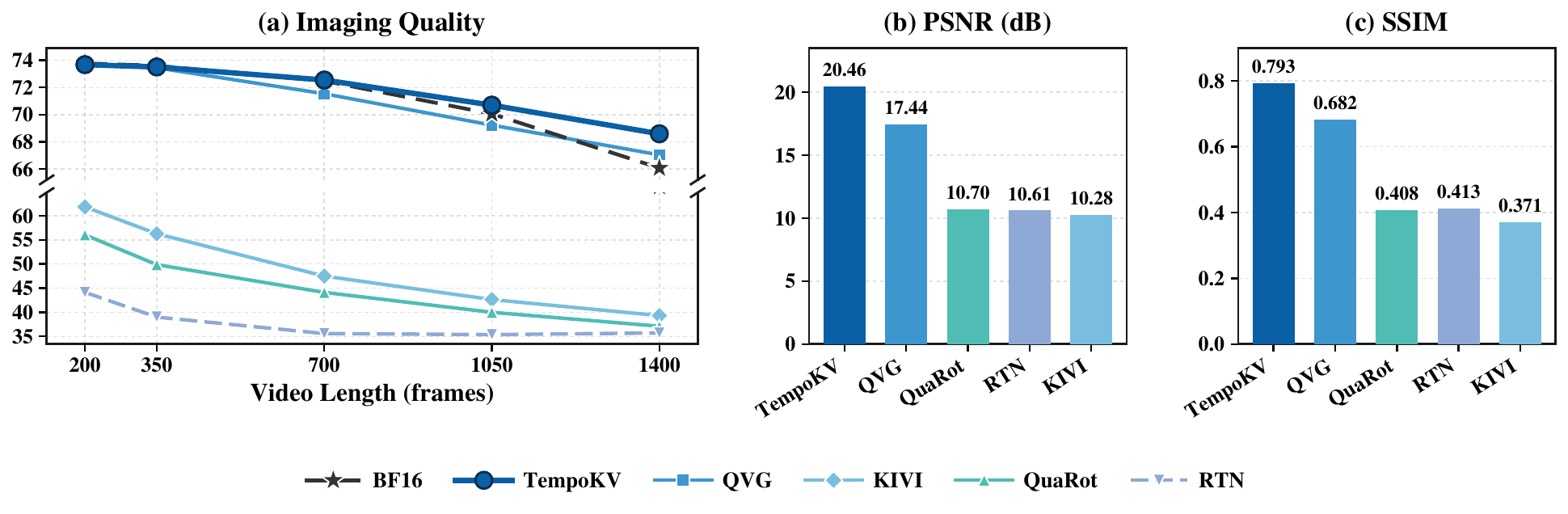}
    \caption{\textbf{Quality preservation over long video generation.}
    (a) Imaging Quality across increasing video lengths, where TempoKV remains close to the BF16 baseline and consistently outperforms existing quantization methods. (b--c) PSNR and SSIM comparisons further show that TempoKV better preserves quality under aggressive KV compression.}

    \label{fig:longvideo}
    \vspace{-1em}
\end{figure*}

\subsection{Long-Horizon Generation}
\label{sec:long_horizon}
We further evaluate long-horizon generation on Self-Forcing by extending the rollout to 1401 frames, corresponding to approximately 90 seconds. With full-history attention, the BF16 baseline reaches only about 815 frames on our hardware before running out of memory, whereas \method can complete the full 1401-frame rollout while retaining the entire KV cache. For the quality comparison in Fig.~\ref{fig:longvideo}, we use the same fixed attention window of 180 latent frames for all methods, including BF16, to avoid differences caused by attention context.
As shown in Fig.~\ref{fig:longvideo}, \method remains closer to the BF16 reference as the generation horizon increases. Compared with QVG, it improves Imaging Quality from 71.54 to \textbf{72.55} at 700 frames, from 69.23 to \textbf{70.70} at 1050 frames, and from 67.04 to \textbf{68.59} at 1400 frames. Averaged across all lengths, \method achieves the highest Imaging Quality among compressed methods, together with \textbf{20.46} PSNR and \textbf{0.793} SSIM.

\subsection{Ablation Studies}
\label{sec:ablation}

\begin{figure}[t]
    \centering
    \vspace{-1em}
    \includegraphics[width=\linewidth]{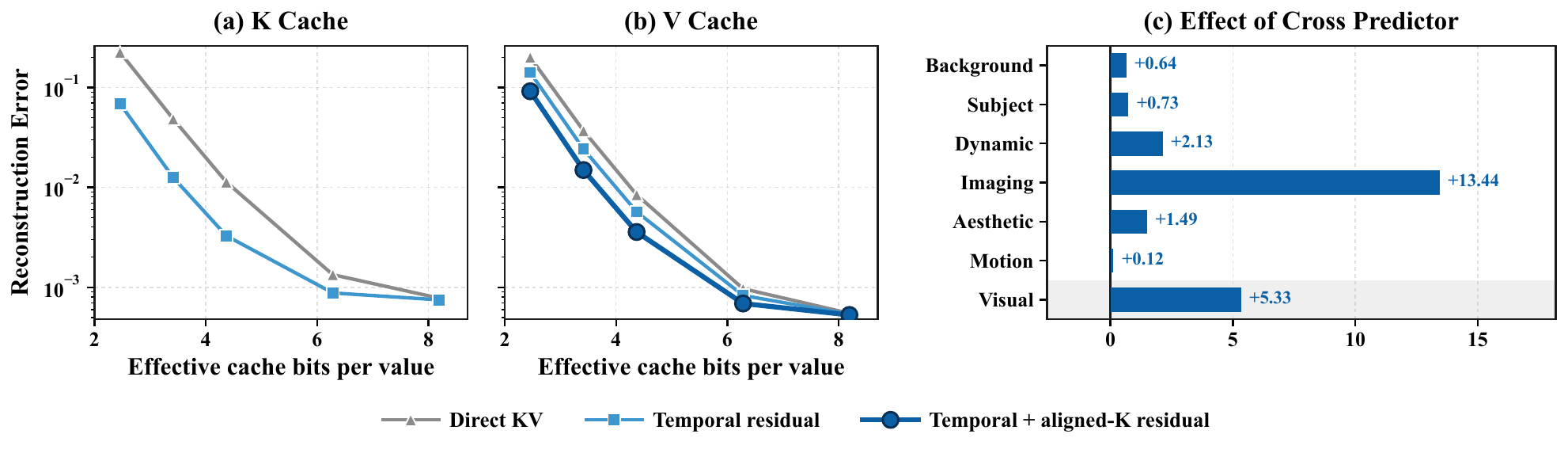}
    \caption{\textbf{Predictor ablation.}
    (a--b) Reconstruction error of the K and V caches on different effective bits. (c) Per-dimension difference between the cross predictor
    and the temporal-only predictor.}
    \label{fig:temporal_rd_v_predictors}
\end{figure}

\paragraph{Effect of the cross predictor.}
Fig.~\ref{fig:temporal_rd_v_predictors}(a--b) compares the normalized reconstruction error at different effective bit-widths. For K, temporal prediction lowers the error at every bit-width, by a factor of $3.27\times$ at INT2.
For V, temporal prediction alone reduces the error by $1.42\times$ while adding the cross predictor raises the reduction to $2.19\times$. The lower error propagates to video quality. Fig.~\ref{fig:temporal_rd_v_predictors}(c) reports the VBench-Long score difference between the cross predictor and the temporal-only predictor.
Visual Quality improves from $68.77$ to $74.10$.
The gain is concentrated in Imaging Quality ($+13.44$), followed by Aesthetic Quality ($+1.49$).
Background Consistency, Subject Consistency and Motion Smoothness change by less than $0.8$.
These three dimensions already exceed $95$ without the cross predictor and are close to saturation.
Consistency scores can also be raised by generating more static video. However, cross predictor's Dynamic Degree increases by $2.13$, indicating that consistency scores were not maintained at the expense of reducing the magnitude of motion.

\begin{table*}[t]
\centering

    \begin{minipage}[b]{0.47\textwidth}
    \centering
    \captionof{table}{
Ablation on mixed-bit K/V precision allocation. Representative settings used in the main experiments
are highlighted.
}
\label{tab:mixed_bits}
\vspace{-1em}

\scriptsize
\setlength{\tabcolsep}{1.8pt}
\renewcommand{\arraystretch}{1.03}

\begin{tabular}{@{}l|c|ccccc@{}}
\toprule
\textbf{Setting}
&
\makecell[c]{\textbf{Compression}\\\textbf{Ratio}}
&
\textbf{PSNR} $\uparrow$
&
\textbf{SSIM} $\uparrow$
&
\textbf{LPIPS} $\downarrow$
&
\textbf{FID} $\downarrow$
&
\makecell[c]{\textbf{Visual}\\\textbf{Quality} $\uparrow$}
\\
\midrule

\rowcolor{blue!8}
\textbf{\texttt{k6v4}}
& 2.97$\times$
& 24.71
& 0.89
& 0.05
& 3.28
& 74.10 \\

\texttt{k5v4}
& 3.28$\times$
& 24.43
& 0.88
& 0.06
& 4.03
& 74.06 \\

\rowcolor{blue!8}
\textbf{\texttt{k5v3}}
& 3.65$\times$
& 24.17
& 0.88
& 0.06
& 4.13
& 74.04 \\

\texttt{k3v5}
& 3.65$\times$
& 22.60
& 0.84
& 0.07
& 6.36
& 73.71 \\

\texttt{k4v4}
& 3.66$\times$
& 23.55
& 0.86
& 0.06
& 4.99
& 73.96 \\

\texttt{k4v3}
& 4.13$\times$
& 23.21
& 0.85
& 0.07
& 5.20
& 73.95 \\

\rowcolor{blue!8}
\textbf{\texttt{k4v2}}
& 4.73$\times$
& 23.17
& 0.85
& 0.07
& 5.23
& 74.04 \\

\texttt{k3v3}
& 4.73$\times$
& 22.47
& 0.84
& 0.07
& 6.43
& 73.92 \\

\texttt{k2v4}
& 4.73$\times$
& 19.83
& 0.78
& 0.13
& 10.28
& 73.58 \\

\texttt{k3v2}
& 5.55$\times$
& 22.22
& 0.83
& 0.08
& 6.94
& 73.75 \\

\texttt{k2v3}
& 5.55$\times$
& 19.77
& 0.78
& 0.13
& 10.36
& 73.64 \\

\rowcolor{blue!8}
\textbf{\texttt{k2v2}}
& 6.71$\times$
& 19.85
& 0.78
& 0.13
& 10.04
& 73.32 \\

\midrule

\texttt{k4v1}
& 5.55$\times$
& 12.31
& 0.45
& 0.61
& 139.63
& 66.09 \\

\texttt{k3v1}
& 6.71$\times$
& 12.33
& 0.45
& 0.62
& 141.82
& 65.71 \\

\bottomrule
\end{tabular}
    \end{minipage}
    \hfill
    \begin{minipage}[b]{0.47\textwidth}
    \centering
    \captionof{table}{
Ablation on the temporal history length for K and V prediction.
}
\label{tab:history_ablation_simple}
\vspace{-1em}

\scriptsize
\setlength{\tabcolsep}{1.8pt}
\renewcommand{\arraystretch}{1.03}

\begin{tabular}{@{}l|c|ccccc@{}}
\toprule
\textbf{Setting}
&
\textbf{$h_K$/$h_V$}
&
\textbf{PSNR} $\uparrow$
&
\textbf{SSIM} $\uparrow$
&
\textbf{LPIPS} $\downarrow$
&
\textbf{FID} $\downarrow$
&
\makecell[c]{\textbf{Visual}\\\textbf{Quality} $\uparrow$}
\\
\midrule

\multicolumn{7}{c}{\textit{Key history (V: w/ aligned-K)}} \\
\midrule

& 0 / 1
& 16.45
& 0.604
& 0.321
& 73.98
& 68.02 \\

& 1 / 1
& 19.73
& 0.776
& 0.134
& 10.42
& \underline{73.06} \\

\rowcolor{blue!8}
\textbf{default}
& \textbf{2 / 1}
& \textbf{19.85}
& \textbf{0.783}
& \textbf{0.130}
& \textbf{10.04}
& \textbf{73.32} \\

& 3 / 1
& \underline{19.81}
& \underline{0.779}
& \underline{0.131}
& \underline{10.11}
& 73.21 \\

& 4 / 1
& 19.76
& 0.777
& 0.133
& 10.28
& 73.24 \\

& 5 / 1
& 19.79
& 0.778
& 0.132
& 10.19
& 73.29 \\

\midrule
\multicolumn{7}{c}{\textit{Value history (V: w/ aligned-K)}} \\
\midrule

& 2 / 0
& 19.73
& 0.775
& 0.135
& 10.51
& 72.93 \\

\rowcolor{blue!8}
\textbf{default}
& \textbf{2 / 1}
& \textbf{19.85}
& \textbf{0.783}
& \textbf{0.130}
& \textbf{10.04}
& \textbf{73.32} \\

& 2 / 2
& \underline{19.80}
& \underline{0.778}
& \underline{0.132}
& \underline{10.16}
& \underline{73.29} \\

& 2 / 3
& 19.77
& 0.777
& 0.133
& 10.25
& 73.15 \\

& 2 / 4
& 19.74
& 0.776
& 0.134
& 10.37
& 73.18 \\

& 2 / 5
& 19.78
& \underline{0.778}
& \underline{0.132}
& 10.21
& 72.99 \\

\bottomrule
\end{tabular}
    \end{minipage}

\end{table*}

\paragraph{Mixed KV bit-widths.}
Sec.~\ref{sec:asymmetry} argues that K should be given higher precision than V. Table~\ref{tab:mixed_bits}  validates this design under different allocation schemes. At $4.73\times$, \textbf{\texttt{k4v2}} substantially outperforms the reversed \textbf{\texttt{k2v4}} allocation, improving PSNR from $19.83$ to \textbf{23.17} and reducing FID from $10.28$ to \textbf{5.23}. 
The same pattern appears at $5.55\times$, where \textbf{\texttt{k3v2}} clearly outperforms \textbf{\texttt{k2v3}}.
The two directions of Table~\ref{tab:mixed_bits} are also asymmetric. Holding K at 2 bits and raising V from 2 to 4 leaves video quality virtually unchanged, while the compression ratio drops from 6.71$\times$ to 4.73$\times$. Holding V at 2 bits and raising K from 2 to 4 improves PSNR by 3.3, from 19.85 to \textbf{23.17}. This further supports allocating additional precision to K rather than distributing the budget uniformly between K and V. 
However, values cannot be assigned excessively low precision. With 1-bit values, the quality collapses sharply.

\paragraph{Temporal history length.}
\label{history ablation}
Table~\ref{tab:history_ablation_simple} studies the temporal history used by the K and V predictors and all settings in this ablation use \textbf{\texttt{k2v2}}. We first vary $h_K$ while fixing $h_V{=}1$ and retaining
aligned-K conditioning for V. Without temporal K prediction ($h_K{=}0$), PSNR drops from $19.85$ to $16.45$ and FID increases from $10.04$ to $73.98$.
Using one previous K frame recovers most of this loss, while $h_K{=}2$ gives the best overall performance. Extending the history to three or more frames
brings no consistent gain across the evaluation metrics. We therefore set $h_K{=}2$ by default. We then vary $h_V$ while fixing $h_K{=}2$ and keeping aligned-K conditioning.
The results are much less sensitive to the V history length. Using one previous V frame gives the best overall performance, while longer histories produce only
minor fluctuations and no consistent improvement. This behavior is consistent with the aligned K providing most of the useful information for V prediction,
with temporal V history serving as a short complementary context. We therefore use $h_V{=}1$ throughout the experiments.

\section{Conclusion}
\label{sec:conclusion}

In this work, we introduce \textbf{\method}, a training-free predictive KV cache quantization framework for autoregressive video generation. 
\method exploits asymmetric dependencies
between keys and values. Keys are predictable from their own history, while values depend mainly on the aligned current key. It predicts both accordingly, quantizes only the residuals, and assigns keys higher precision. 
Extensive experiments demonstrate that \method substantially improves low-bit KV reconstruction and achieves up to $6.71\times$ cache compression, while delivering the best generation quality and compression--quality trade-off compared with existing KV cache quantization work.

\newpage

\bibliography{iclr2027_conference}
\bibliographystyle{iclr2027_conference}

\newpage
\appendix

\appendix

\section{Additional Experiments}

\subsection{Predictability on LongCat-Video-13B}
\label{app:longcat_cosine}
\begin{figure}[H]
    \centering
    \includegraphics[width=\textwidth]{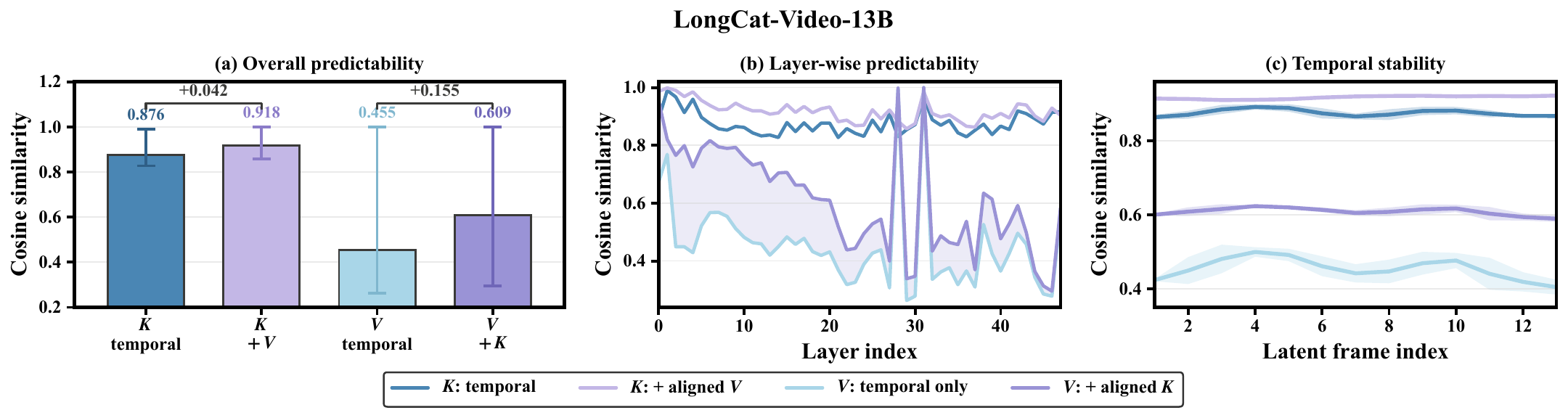}
    \caption{Predictability of the key and value caches on LongCat-Video-13B, measured by cosine similarity between predicted and original caches.
    (a) Mean over all layers, with and without the aligned cache as input. (b) Per layer. (c) Per latent frame. }
    \label{fig:longcat-cosine}
\end{figure}

Fig.~\ref{fig:longcat-cosine} repeats the predictability analysis on LongCat-Video-13B using cosine similarity between the predicted and original caches. The same K--V asymmetry remains clearly visible on this larger backbone: adding the aligned V improves K prediction from 0.876 to 0.918 (\textbf{+0.042}), whereas adding the aligned K improves V prediction from 0.455 to 0.609 (\textbf{+0.155}), yielding a much larger gain.
 
Predictability varies substantially across layers but remains stable over time. With temporal prediction alone, the layer-wise cosine similarity ranges from 0.827 to 0.990 for K and from 0.263 to nearly 1.0 for V. Adding the aligned cache preserves this large layer-wise variation. In contrast, the variation across latent frames is much smaller: \emph{K + aligned V} remains between 0.909 and 0.924, while \emph{V + aligned K} stays between 0.581 and 0.625. Across all four predictors, the layer-wise variation is approximately $3\times$--$13\times$ larger than the temporal variation. This layer-dependent but temporally stable behavior supports the design of \method: we fit a separate predictor for each layer and head and reuse it across all latent frames. These results show that the asymmetric predictive dependencies identified in Sec.~\ref{sec:asymmetry} persist across different autoregressive video models.
 
\subsection{Cross-prediction direction}
We further test whether the cross prediction between K and V depends on its direction by reversing the predictor from K$\to$V to V$\to$K, with the rest of the
compression pipeline unchanged. As shown in Table~\ref{tab:cross_direction}, K$\to$V achieves better reconstruction quality across all bit allocations. 
The gap is particularly large at low precision. Under \textbf{\texttt{k4v2}}, reversing the direction reduces PSNR from \textbf{23.17} to \textbf{17.71} and
increases FID from \textbf{5.23} to \textbf{73.80}. 
The difference remains at higher precision, e.g. \textbf{\texttt{k5v3}} and \textbf{\texttt{k6v4}}.
These results indicate that the aligned key is a better predictor of values, which supports the
K$\to$V direction and the asymmetric bit allocation in \method.
\begin{table}[H]
  \centering
  \small
  \caption{\textbf{Cross predictor is directional.} Predicting V from the
    reconstructed K outperforms the reverse at every bit allocation and on every
    metric. The better direction is shown in \textbf{bold}.}
  \label{tab:cross_direction}
  \begin{tabular}{@{}llcccc@{}}
    \toprule
    Bits (K/V) & Direction & PSNR$\uparrow$ & SSIM$\uparrow$ & LPIPS$\downarrow$ & FID$\downarrow$\\
    \midrule
    \rowcolor{blue!8}
    \cellcolor{white}\multirow{2}{*}{2 / 2}
        & K$\to$V & \textbf{19.85} & \textbf{0.7822} & \textbf{0.1303} & \textbf{10.04}\\
                           & V$\to$K & 17.66 & 0.6394 & 0.2575 & 75.73\\
    \midrule
    \rowcolor{blue!8}
    \cellcolor{white}\multirow{2}{*}{4 / 2} 
    & K$\to$V & \textbf{23.17} & \textbf{0.8539} & \textbf{0.0729} & \textbf{5.23}\\
                           & V$\to$K & 17.71 & 0.6426 & 0.2562 & 73.80\\
    \midrule
    \rowcolor{blue!8}
    \cellcolor{white}\multirow{2}{*}{5 / 3} 
    & K$\to$V & \textbf{24.17} & \textbf{0.8815} & \textbf{0.0597} & \textbf{4.13}\\
                           & V$\to$K & 20.74 & 0.7886 & 0.1139 & 17.24\\
    \midrule
    \rowcolor{blue!8}
    \cellcolor{white}\multirow{2}{*}{6 / 4} 
    & K$\to$V & \textbf{24.71} & \textbf{0.8947} & \textbf{0.0526} & \textbf{3.28}\\
                           & V$\to$K & 22.60 & 0.8436 & 0.0763 & 7.62\\
    \bottomrule
  \end{tabular}
\end{table}

\subsection{Anchor precision}
Table~\ref{tab:anchor_ablation} studies the anchor precision under \textbf{\texttt{k2v2}}. Anchors below 4 bits degrade reconstruction quality. FID rises from 10.04
with 4-bit anchors to 23.31 and 65.88 with 3-bit and 2-bit anchors, respectively. Above 4 bits, both reconstruction and VBench scores remain largely unchanged, while the compression ratio decreases as more
bits are allocated to anchors. Therefore, we set
$b_a=\max(4,b_K,b_V)$, which keeps anchors at least 4-bit while providing a sufficiently accurate reference for subsequent prediction without unnecessary storage overhead.
\begin{table}[H]
\centering
\small
\vspace{-1em}
\caption{\textbf{Ablation on anchor precision.}
Anchor precisions below 4 bits degrade reconstruction quality, whereas increasing the precision beyond 4 bits provides little additional benefit.}
\label{tab:anchor_ablation}

\resizebox{\textwidth}{!}{%

\begin{tabular}{c|c|cccc|cccc}
\toprule
\multirow{2}{*}{\textbf{Anchor bits}}
& \multirow{2}{*}{\textbf{Comp.$\uparrow$}}
& \multirow{2}{*}{\textbf{PSNR$\uparrow$}}
& \multirow{2}{*}{\textbf{SSIM$\uparrow$}}
& \multirow{2}{*}{\textbf{LPIPS$\downarrow$}}
& \multirow{2}{*}{\textbf{FID$\downarrow$}}
& \textbf{Subject}
& \textbf{Motion}
& \textbf{Aesthetic}
& \textbf{Imaging} \\
&&&&&&
\textbf{Consistency$\uparrow$}
& \textbf{Smoothness$\uparrow$}
& \textbf{Quality$\uparrow$}
& \textbf{Quality$\uparrow$} \\
\midrule

2
& 6.739$\times$
& 16.83 & 0.6095 & 0.2742 & 65.88
& 96.68 & 98.28 & 39.30 & 62.61 \\

3
& 6.723$\times$
& 19.53 & 0.7304 & 0.1471 & 23.31
& 96.96 & 98.43 & 39.69 & 67.35 \\

\rowcolor{blue!8}
\textbf{4}
& \textbf{6.707$\times$}
& \textbf{19.85} & \textbf{0.7832} & \textbf{0.1303} & \textbf{10.04}
& \textbf{97.04} & \textbf{98.48} & \textbf{39.79} & \textbf{67.81} \\

5
& 6.692$\times$
& 19.79 & 0.7858 & 0.1243 & 10.09
& 97.04 & 98.49 & 39.83 & 68.02 \\

6
& 6.676$\times$
& 19.28 & 0.7818 & 0.1244 & 10.12
& 97.04 & 98.50 & 39.79 & 67.87 \\

7
& 6.661$\times$
& 19.81 & 0.7841 & 0.1242 & 10.01
& 97.06 & 98.49 & 39.78 & 67.73 \\

8
& 6.646$\times$
& 19.67 & 0.7808 & 0.1270 & 10.04
& 97.08 & 98.49 & 39.84 & 68.13 \\

16 (BF16)
& 6.529$\times$
& 19.57 & 0.7797 & 0.1274 & 10.07
& 97.04 & 98.50 & 39.83 & 67.84 \\

\bottomrule

\end{tabular}%
}
\end{table}

\subsection{Higher-Resolution Generation}
To evaluate the effectiveness of \method at higher resolutions, we conduct experiments on LongCat-Video-13B at 720p under INT4 quantization. As shown in Table~\ref{tab:higher-resolutions}, \method\textbf{\texttt{k5v3}} achieves the highest compression ratio of $3.65\times$ while also providing the best reconstruction fidelity, with a PSNR of $17.31$, SSIM of $0.77$, LPIPS of $0.12$, and FID of
$11.20$. It also attains the highest Visual Quality score of $75.74$ and the best Subject Consistency ($97.76$), Dynamic Degree ($47.33$), Imaging
Quality ($74.64$), and Motion Smoothness ($98.37$) among the compression methods.
These results show that the predictive structure exploited by \method remains effective when the number of spatial tokens increases, preserving its compression advantage and generation quality at higher resolutions.
\begin{table*}[!ht]
    \centering
    \caption{Quality comparison on LongCat-Video at 720p under INT4 quantization. Best and second-best results are shown in \textbf{bold} and \underline{underline}, respectively.}
    \vspace{-0.3em}
    \label{tab:higher-resolutions}
    \resizebox{\textwidth}{!}{
    \begin{tabular}{l|c|cccc|cccccc|c}
        \toprule
        \textbf{Method}
        & \textbf{Compression}
        & \textbf{PSNR$\uparrow$}
        & \textbf{SSIM$\uparrow$}
        & \textbf{LPIPS$\downarrow$}
        & \textbf{FID$\downarrow$}
        & \makecell{\textbf{Background}\\\textbf{Consistency$\uparrow$}}
        & \makecell{\textbf{Subject}\\\textbf{Consistency$\uparrow$}}
        & \makecell{\textbf{Dynamic}\\\textbf{Degree$\uparrow$}}
        & \makecell{\textbf{Imaging}\\\textbf{Quality$\uparrow$}}
        & \makecell{\textbf{Aesthetic}\\\textbf{Quality$\uparrow$}}
        & \makecell{\textbf{Motion}\\\textbf{Smoothness$\uparrow$}}
        & \makecell{\textbf{Visual}\\\textbf{Quality$\uparrow$}} \\
        \midrule

        BF16
        & 1.00$\times$
        & $\infty$
        & 1.00
        & 0.00
        & 0.00
        & 95.68
        & 97.71
        & 47.67
        & 74.71
        & 41.60
        & 98.29
        & 76.28 \\
        \midrule

        RTN
        & 3.20$\times$
        & 15.65
        & 0.72
        & 0.17
        & 15.23
        & 95.28
        & 97.61
        & 43.12
        & 74.33
        & 40.59
        & 98.32
        & 74.87 \\

        KIVI
        & 3.20$\times$
        & 13.13
        & 0.61
        & 0.32
        & 59.83
        & 95.60
        & 94.81
        & 26.67
        & 65.02
        & 39.14
        & 96.39
        & 69.61 \\

        QuaRot
        & 3.20$\times$
        & \underline{16.48}
        & \underline{0.74}
        & \underline{0.15}
        & \underline{13.56}
        & 95.73
        & \underline{97.72}
        & 45.67
        & 74.32
        & 40.58
        & \underline{98.33}
        & 75.40 \\

        QVG
        & \underline{3.59$\times$}
        & 15.43
        & 0.66
        & 0.23
        & 18.94
        & 95.58
        & 97.51
        & \underline{47.00}
        & 74.32
        & 40.60
        & 98.18
        & \underline{75.53} \\

        QVG-Pro
        & 2.75$\times$
        & 15.32
        & 0.68
        & 0.20
        & 19.60
        & \textbf{95.82}
        & 97.52
        & 46.00
        & \underline{74.57}
        & \textbf{40.63}
        & 98.28
        & 75.47 \\

        \rowcolor{blue!8}
        \textbf{TempoKV (\texttt{k5v3})}
        & \textbf{3.65$\times$}
        & \textbf{17.31}
        & \textbf{0.77}
        & \textbf{0.12}
        & \textbf{11.20}
        & \underline{95.73}
        & \textbf{97.76}
        & \textbf{47.33}
        & \textbf{74.64}
        & \underline{40.62}
        & \textbf{98.37}
        & \textbf{75.74} \\

        \bottomrule
    \end{tabular}
    
    }
    \vspace{-1em}
\end{table*}

\subsection{Calibration Analysis}
\begin{figure}[H]
    \centering

    \begin{minipage}[t]{0.47\textwidth}
        \centering
        \vspace{0pt}
        \includegraphics[
            width=\linewidth,
        ]{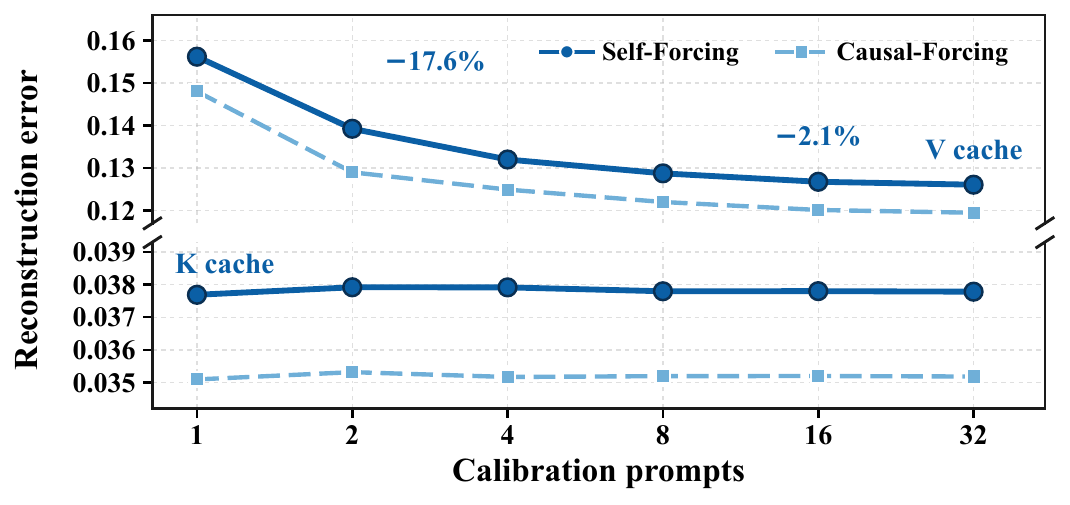}

        \vspace{2pt}
        {\small (a) Few calibration prompts are sufficient}
    \end{minipage}
    \hfill
    \begin{minipage}[t]{0.50\textwidth}
        \centering
        \vspace{8pt}

        \footnotesize
        \setlength{\tabcolsep}{6pt}
        \renewcommand{\arraystretch}{1.12}

        \resizebox{\linewidth}{!}{%
        \begin{tabular}{@{}lcccc@{}}
            \toprule
            \diagbox{\textbf{Calib.}}{\textbf{Eval}}
            & \multicolumn{2}{c}{\textbf{Action}}
            & \multicolumn{2}{c}{\textbf{Scene}} \\
            \cmidrule(lr){2-3}
            \cmidrule(lr){4-5}
            & \textbf{K} & \textbf{V}
            & \textbf{K} & \textbf{V} \\
            \midrule
            Action
            & \textbf{0.03734}
            & \textbf{0.12644}
            & 0.03828
            & 0.12660 \\

            Scene
            & 0.03730
            & 0.12742
            & \textbf{0.03823}
            & \textbf{0.12744} \\
            \midrule

            $\Delta$ vs.\ matched
            & $-0.11\%$
            & $+0.77\%$
            & $+0.11\%$
            & $-0.66\%$ \\
            \bottomrule
        \end{tabular}%
        }

        \vspace{10pt}
        {\small (b) Robust to calibration-set distribution}
    \end{minipage}

    \caption{\textbf{Calibration of TempoKV.}
    \textbf{Left:} A small calibration set is sufficient for TempoKV. Reconstruction error drops rapidly with the first few calibration prompts and nearly saturates thereafter on both Self-Forcing and Causal Forcing, with little additional gain beyond a handful of prompts.
    \textbf{Right:} TempoKV is robust to calibration-set distribution. Switching between action and scene prompts changes reconstruction error by at most $0.77\%$ relative to the matched setting.}
    \label{fig:calibration}
\end{figure}

Fig.~\ref{fig:calibration} studies how many calibration prompts the predictors need and how sensitive they are to the calibration distribution. We build
two disjoint calibration sets with different content. The \emph{action} set focuses on motion and interactions, and the \emph{scene} set focuses on appearance, composition, and environment. No calibration prompt overlaps with the MovieGen prompts used for evaluation.

\paragraph{A few calibration prompts are sufficient.}
We first vary the number of calibration prompts from 1 to 32. The K predictor is particularly data-efficient: a single prompt already achieves a reconstruction error within $0.3\%$ of that obtained with 32
prompts. The cross predictor benefits more from additional calibration data, with its error decreasing by $17.6\%$ from 1 to 8 prompts. Beyond eight prompts, however, the improvement quickly saturates, with only a further $2.1\%$ reduction from 8 to 32 prompts. We use 8 calibration prompts by default to collect the BF16 traces for predictor fitting. These results show that a small calibration set is sufficient to obtain stable predictors.

\paragraph{Robust to calibration-set distribution.}
We next calibrate the predictors separately on the action and scene sets and evaluate each predictor on both distributions. Using a mismatched calibration
set changes the reconstruction error by at most $0.77\%$ across K and V. Moreover, the change is not consistently negative: cross-distribution calibration slightly improves some settings while slightly degrading others.
The predictors are therefore largely insensitive to the content distribution of the calibration set. Together with the small calibration requirement above, this makes \method lightweight to calibrate and removes the need for a large or carefully matched calibration corpus.

\subsection{Comparison with token pruning}
\label{app:pruning}
We compare \method with Forcing-KV~\citep{ji2026forcingkv}, which reduces KV cache memory by pruning cached tokens.
Table~\ref{tab:forcingkv_comparison} reports generation quality.
On Causal Forcing, Forcing-KV reaches a PSNR of 9.32 and an FID of 40.07, while \method in its lowest-bit configuration reaches 17.23 and 9.15. The gap is
larger on Self-Forcing, where Forcing-KV reaches 9.44 and 83.86 and \method reaches 19.85 and 10.04.
Higher-bit configurations of \method improve fidelity further, and all \method configurations score higher than Forcing-KV in Visual Quality. 
Compared with token pruning, quantization preserves more information from each cached token and yields higher fidelity. Token pruning, in turn, reduces attention computation and can accelerate generation.
\begin{table}[H]
  \centering
  \vspace{-1em}
  \small
    \caption{\textbf{Comparison with token pruning.}
    Across both Causal Forcing and Self-Forcing, \method improves fidelity in PSNR, SSIM, LPIPS, and FID, while also achieving higher overall Visual Quality.}
  \label{tab:forcingkv_comparison}
  \setlength{\tabcolsep}{7pt}
  \begin{tabular}{l|cccc|c}
    \toprule
    \textbf{Method}
    & \textbf{PSNR$\uparrow$}
    & \textbf{SSIM$\uparrow$}
    & \textbf{LPIPS$\downarrow$}
    & \textbf{FID$\downarrow$}
    & \textbf{Visual Quality$\uparrow$} \\
    \midrule

    \multicolumn{6}{c}{\emph{Causal Forcing, 480p}} \\
    \midrule
    Forcing-KV
    & 9.32 & 0.38 & 0.58 & 40.07 & 76.70 \\
    
    \textbf{TempoKV\texttt{(k2v2)}}
    & 17.23 & 0.76 & 0.17 & 9.15 & 77.04 \\

    \textbf{TempoKV\texttt{(k4v2)}}
    & 22.85 & 0.79 & 0.13
    & 6.34 & 77.11 \\
    
    \textbf{TempoKV\texttt{(k5v3)}}
    & \textbf{23.21} & \textbf{0.80} & \textbf{0.11}
    & \textbf{4.03} & \textbf{77.12} \\

    \midrule
    \multicolumn{6}{c}{\emph{Self-Forcing, 480p}} \\
    \midrule
    Forcing-KV
    & 9.44 & 0.25 & 0.71 & 83.86 & 72.79 \\
    
    \textbf{TempoKV\texttt{(k2v2)}}
    & 19.85 & 0.78 & 0.13 & 10.04 & 73.32 \\
    
    \textbf{TempoKV\texttt{(k4v2)}}
    & 23.17 & 0.85 & 0.07 & 5.23 & 74.04 \\
    
    \textbf{TempoKV\texttt{(k5v3)}}
    & \textbf{24.17} & \textbf{0.88} 
    & \textbf{0.06} & \textbf{4.13} 
    & \textbf{74.04} \\
    
    \bottomrule
  \end{tabular}
\end{table}

\section{Visual Quality Comparisons}


\begin{figure*}[!ht]
    \centering
    \vspace{-1em}
    \includegraphics[
        width=\textwidth
    ]{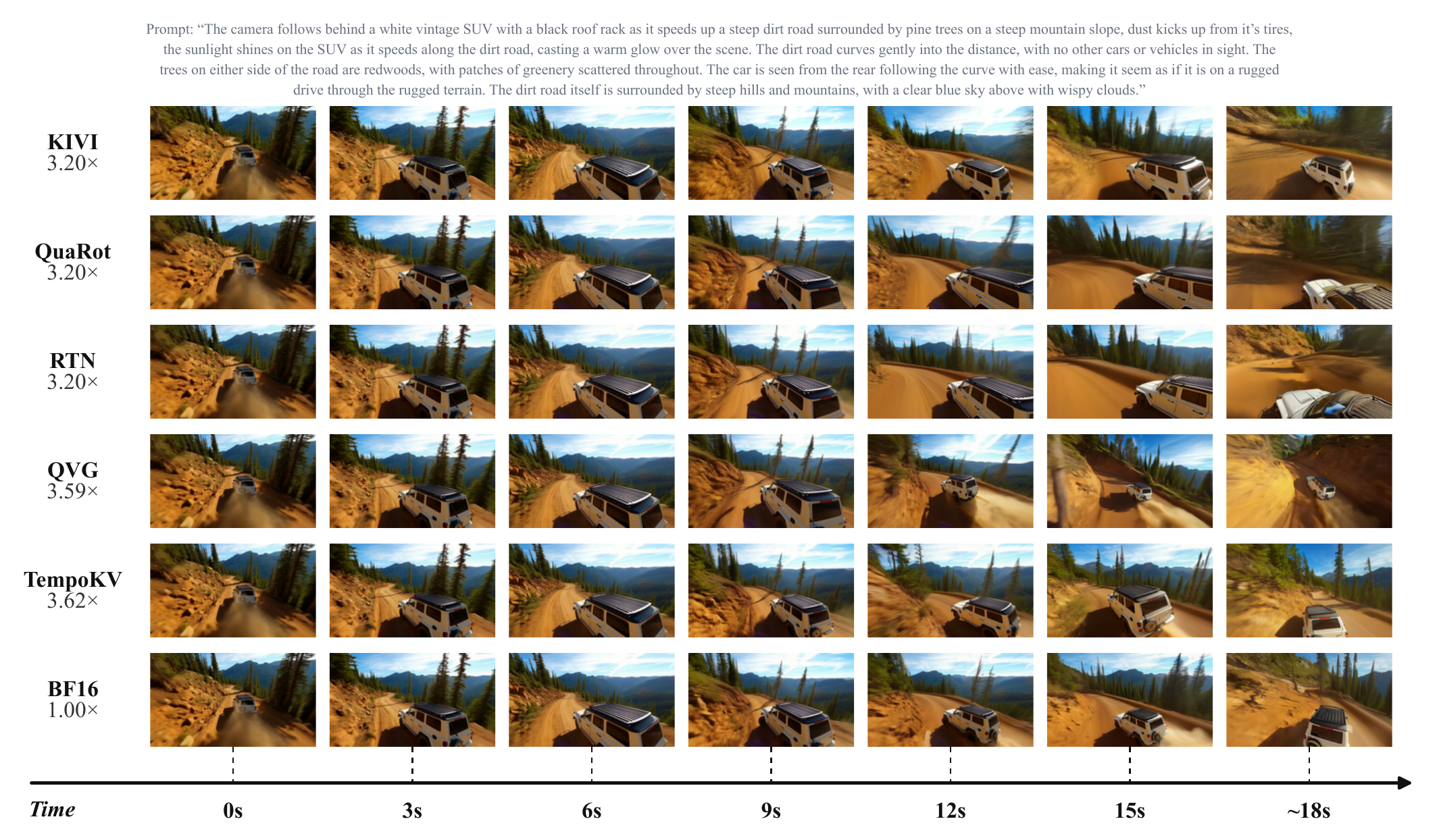}
    \caption{Qualitative comparison on LongCat-Video-13B at INT4. TempoKV maintains superior temporal coherence and visual fidelity at a higher compression ratio.}
    \label{fig:longcat_comparison}
\end{figure*}

\begin{figure*}[!ht]
    \centering
    \vspace{-1em}
    \includegraphics[
        width=\textwidth
    ]{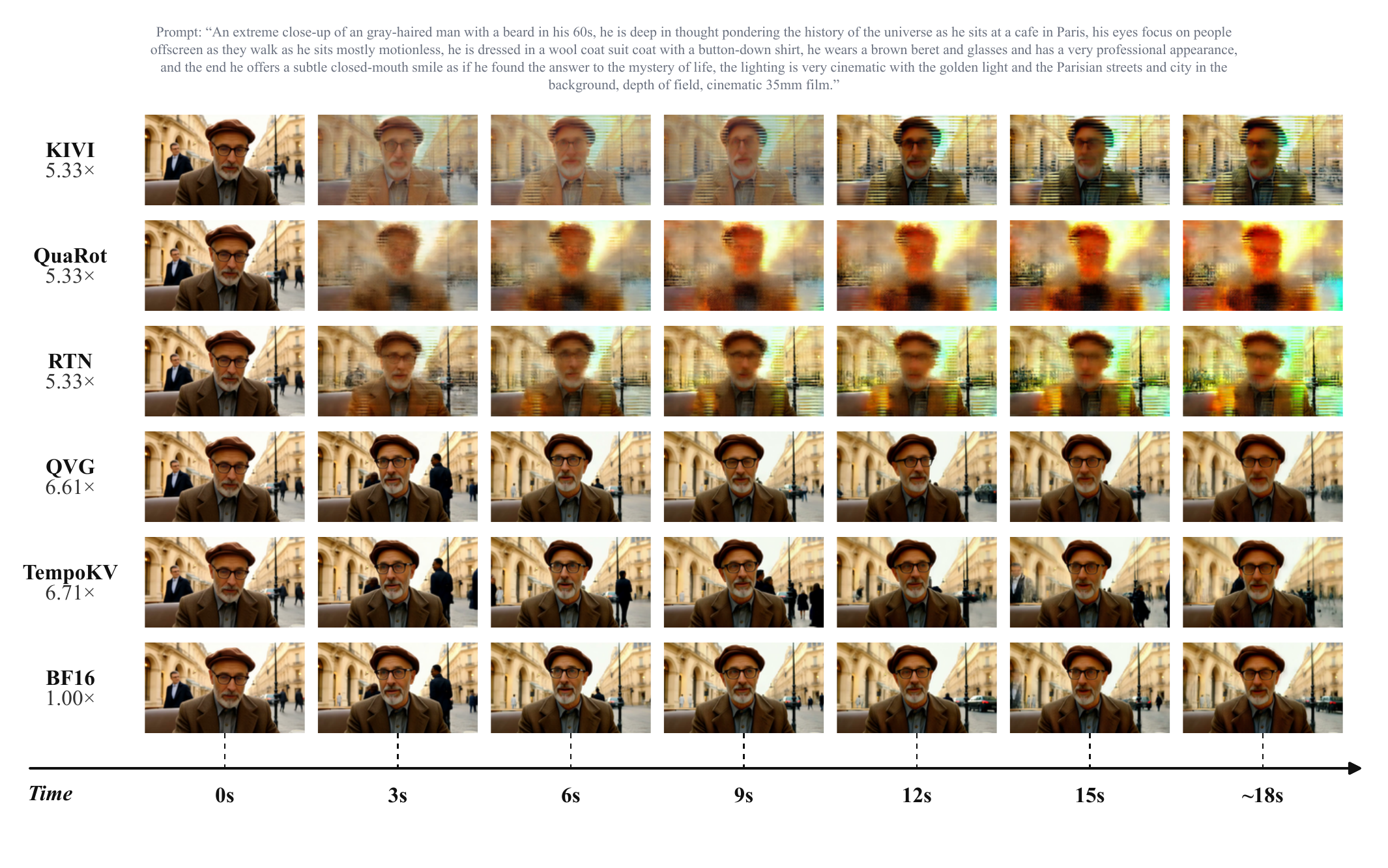}
    \caption{Qualitative comparison on Causal Forcing at INT2. TempoKV maintains superior temporal coherence and visual fidelity at a higher compression ratio.}
    \label{fig:longcat_comparison}
    \vspace{-2em}
\end{figure*}

\end{document}